%% file: Main.tex
\documentclass{article}

\usepackage{microtype}
\usepackage{graphicx}
\usepackage{subfigure}
\usepackage{booktabs} 
\usepackage[most]{tcolorbox}

\usepackage{hyperref}

\usepackage[accepted]{icml2024}

\usepackage{amsmath}
\usepackage{amssymb}
\usepackage{mathtools}
\usepackage{amsthm}

\usepackage[capitalize,noabbrev]{cleveref}

\theoremstyle{plain}

\theoremstyle{definition}

\theoremstyle{remark}

\usepackage[textsize=tiny]{todonotes}
\usepackage{hyperref}

\icmltitlerunning{Beyond Simple Input-Output Assessment Tasks: Leveraging Automated Programming Assessment for Non-Trivial Courses}

\begin{document}

\twocolumn[
\icmltitle{Beyond Simple Input-Output Assessment Tasks:\\Leveraging Automated Programming Assessment for Non-Trivial Courses}



\icmlsetsymbol{equal}{*}

\begin{icmlauthorlist}
\icmlauthor{Artur Jordao}{usp}
\end{icmlauthorlist}

%

\vskip 0.3in
]





\begin{abstract}
The public visibility of Artificial Intelligence (AI) is growing rapidly, driven by the positive impact of its applications across diverse fields of knowledge. In this new chapter, courses that cover the foundations of AI and machine learning become essential for understanding their role and potential in contemporary society. 
Therefore, understanding fundamental concepts and elementary algorithms through the close integration of theory with practice is essential in AI courses. In this essay, we report our experience designing machine learning exercises for automated assessment tools in programming. It is worth mentioning that we are not developing a novel form of automated grading system. Instead, we propose a perspective that frames machine learning problems as input-output assessment tasks. From this perspective, each exercise admits a \emph{unique and deterministic answer} and enables automated programming assessment tools (e.g., VPL for Moodle, Codeforces, and MOJ) to effectively support AI education. We believe this essay can encourage instructors to foster educational innovation by adopting more dynamic and interactive approaches to AI courses that integrate theory and practice. Importantly, this essay \textbf{\emph{does not introduce an innovation in the use of AI for education}}; rather, it introduces an \textbf{\emph{innovative approach to improving the learning of AI}}, particularly, machine learning.
\end{abstract}

\input{teaser}
\section{Introduction}
Artificial Intelligence (AI) is gaining greater public visibility, driven in particular by the positive impact of generative AI across diverse fields~\cite{Bengio:2025,Maslej:2025}. As a concrete example, AI tools improve workplace efficiency by automating costly tasks, reducing operational costs, and increasing productivity~\cite{Acqua:2023}. The fields of teaching and education also experience broader adoption of AI~\cite{Maslej:2025}.

Given the current AI wave and its extensive integration across emerging technologies, courses covering the foundations of artificial intelligence, machine learning, and deep learning become increasingly sought-after and essential for understanding the role and potential of AI in contemporary society~\cite{Maslej:2025}. For example, the report by Maslej et al.~\cite{Maslej:2025} found that the number of U.S. institutions offering a dedicated bachelor's degree in AI nearly doubled between 2022 and 2023. The number of institutions offering dedicated master's degrees in AI also grew sharply. In this context, hands-on practice through exercises reinforces the underlying concepts while enhancing analytical and problem-solving skills in AI. 
Additionally, closely aligning theory with practice becomes essential for formulating algorithmic solutions and implementing them through programming languages.

In this essay, we report our experience designing machine learning exercises for automated programming assessment tools.
We leverage the renowned International Collegiate Programming Contest (ICPC) as inspiration to formulate practical exercises covering specific machine learning fundamentals~\cite{icpc}. For AI practitioners and instructors, a natural question is: \emph{why not host a competition on the Kaggle platform}? It turns out that Kaggle competitions allow students (participants) to solve the same problem or task using different machine learning concepts while focusing primarily on the final evaluation metric (i.e., accuracy). Here, on the other hand, we design exercises covering specific machine learning topics, each admitting a \textbf{unique and deterministic answer}. For example, estimating the explained variance captured by the first principal components of PCA or identifying the support vectors given an SVM hyperplane.
For this purpose, we transform machine learning fundamentals into input-output assessment tasks suitable for automated programming assessment tools. Figure~\ref{fig:ExerciseExamples} illustrates an overview of this idea. Powered by these tools, our approach offers clear advantages, including immediate feedback for students and reduced workload for instructors and teaching assistants. We highlight that this input-output assessment strategy is not new and is already common in introductory courses on programming, algorithms and data structures. Our contribution lies in extending this paradigm to machine learning courses.

At the time of writing, we release the exercises through Virtual Programming Lab (VPL) for Moodle, but the same exercises can also run on other automated programming assessment platforms, such as MOJ, Codeforces, or URI Online Judge. Overall, as we shall see, we had a successful experience with this practice, with students providing positive and enthusiastic feedback and largely agreeing that the exercises effectively integrate theory and practice.

\section{Methodology}

\noindent
\textbf{Machine Learning Concepts to Input-Output Format.}
The key to successfully employing automated programming assessment platforms lies in effectively translating machine learning fundamentals into an input-output format. Figure~\ref{fig:ExerciseExamples} introduces two examples of exercises we created for SVM and PCA. Following these examples, we also include exercises on \emph{k-nearest neighbors classifier, ordinary least squares, hierarchical clustering, decision tree, z-score} and \emph{min-max normalization}. 

Due to environment constraints (i.e., VPL), not all techniques support execution from scratch. Specifically, while the environment we use in 2026 supports SVD through \texttt{NumPy}, it does not support quadratic programming or libraries such as \texttt{scikit-learn}, thus preventing students from obtaining some values from scratch (i.e., through a training process) such as the SVM hyperplane. 
In these cases, we adopt the following strategy. First, we run the technique locally in an environment that provides all the required dependencies and compute the necessary values. We then use these values as inputs to the exercise. For example, for SVM, we learn the hyperplane parameters $w$ and $b$ locally and provide these values as inputs to the exercise (see Figure~\ref{fig:ExerciseExamples} left).

\noindent
\textbf{Exercise Writing.}
When translating a concept into an input-output format and designing an exercise, we must consider aspects beyond the core objective to clearly specify how students should solve the problem and avoid cases where different valid approaches produce incompatible outputs. It turns out that some machine learning techniques admit multiple valid solutions, such as computing PCA using either SVD or eigendecomposition of the covariance matrix.
Keep in mind that we are discussing deterministic input-output execution on automated programming assessment platforms.

The naive solution to the previous issue is to explicitly define how students should solve the problem, e.g., \emph{``To compute the SVD, use np.linalg.svd(., full\_matrices=False)''}. The final ingredient when writing an exercise is to address numerical precision. In a similar vein, we explicitly define the required numerical precision for specific calculations or outputs, e.g., \emph{``use the np.isclose(., atol=1e-3) function provided in the code''}.

\noindent
\textbf{On the Importance of Determinism and the Role of Pseudo-random Numbers.}
Many machine learning techniques have a stochastic nature. For example, k-means clustering uses random initial centroids, while random forests rely on feature and sample subsampling, among other sources of randomness. Since automated programming assessment tools require deterministic outputs, we must ensure algorithms produce deterministic results. Concretely, executing a machine learning algorithm on a given set of inputs must produce the same output.

Fortunately, most Python packages involving randomness provide a mechanism for controlling the pseudo-random number generator through a seed, ensuring deterministic (reproducible) results. Therefore, to address the previous issue, we leverage this feature and set the random seed for the packages involved in each exercise (as we shall see, we restrict what packages a student can use). Two options emerge here: (i) provide the seed as a fixed value in the code template, or (ii) provide the seed as an additional input (see the code in Figure~\ref{fig:ExerciseExamples} right). We opt for the latter to ensure that students do not inadvertently modify the seed.

Another important role of pseudo-random numbers is to simplify the test cases in exercise definitions. It turns out that, describing input-output examples for many machine learning concepts can become laborious and even hinder the understanding of students. By leveraging pseudo-random numbers, we can easily generate large, representative examples by providing a seed instead of specifying each number individually. For example, providing a seed that generates huge matrices representing the independent (data) and dependent (label) variables, along with the expected behavior. Figure~\ref{fig:ExerciseExamples} (right) illustrates this idea.

\textbf{On the Importance of Providing a Template Code.}
Since we focus on practicing deep concepts rather than basic programming, algorithms, or data structures, we provide a code template for each exercise to minimize the programming overhead for students. In addition, the code template helps prevent students from using unauthorized solutions and guides them in specifying how to solve the problem.

In a code template, we often provide instructions for reading the inputs (including the seed), importing the required packages, and handling numerical precision when producing the output.
Depending on the complexity of the exercise, we also specify which sections of the code or classes students must complete. The code below provides an overview of a general code template.

{
\small
\begin{lstlisting}[language=Python,
basicstyle=\ttfamily,
keywordstyle=\bfseries\color{blue}
]
import numpy as np
import random

s, n, m = map(int, input().split())
np.random.seed(s)
random.seed(s)

X = np.random.randn(n, m)
\end{lstlisting}
}

\noindent
\textbf{Large Language Models to Mitigate Unauthorized Solutions - SAGE.}
While the code template provides a first line of defense against unauthorized solutions, it is clearly not sufficient on its own. Specifically, a student could employ libraries or packages that would easily solve a problem, thus compromising the goal of practicing the concepts targeted by the exercise. A naive approach to this issue involves manually inspecting each submission for potential violations or unauthorized solutions. Unfortunately, even in medium-size classes, manually checking every submission is infeasible.

Given the success of Large Language Models (LLMs) in coding and recent LLM-as-a-Judge benchmarks~\cite{qwen3,Zheng:2023}, to address the previous challenge, we developed SAGE: an LLM-based system for additional evaluation of student solutions. The idea behind SAGE consists of reading both the exercise statement and the submitted code of a student and assigning a score that reflects whether the student respected all the constraints specified in the problem. From this score, the instructor can identify which student submissions require inspection outside the automated programming assessment platform.

SAGE operates independently of the automated programming assessment platform, running locally to analyze student submissions. We initially developed SAGE using the Gemini API, but its current version employs LM Studio\footnote{https://lmstudio.ai/}. We believe this approach facilitates broader adoption of SAGE, as LM Studio supports a wide range of models and enables fully local, cost-free inference. SAGE is available for download at this \href{https://drive.google.com/drive/folders/12cbx6A1TvrBa4zAtbC_JIsNKeD1vE9FT?usp=sharing}{URL}

To help LLMs identify the constraints and what students must respect, we explicitly state these requirements in a \emph{``Note:''} section at the end of the exercise statement. Figure~\ref{fig:prompt_sage} illustrates an example of prompt of SAGE.

\begin{figure}[!t]
{
\scriptsize
\begin{tcolorbox}[width=0.48\textwidth, colback=white,colframe=black, title=Prompt for SAGE]
[System]
Evaluate the student code against the question statement. Assign an integer score from 0 to 10, considering exclusively how correctly and exactly the code fulfills what the statement requested, including the required format.
Consider:
10: fulfills the statement and the requested format completely.
0: does not fulfill what was requested or presents an essentially incorrect solution.
Intermediate values: partially fulfill the statement or present errors that compromise part of the solution. Do not penalize the student for aspects that the statement did not explicitly request. Your response MUST be only an integer number between 0 and 10, without additional text, explanations, or punctuation.

Mandatory format: \texttt{\textbackslash b(?:10|[0-9])\textbackslash b}

Example of a valid response: 8

\vspace{5pt}

[The Start of Exercise Statement]

\vspace{5pt}

\{Exercise Statement\}

\vspace{5pt}

[The Start of Code Submitted]

\{Code Submitted\}
\end{tcolorbox}
}
\caption{SAGE default prompt. We fill the markers \texttt{[The Start of Exercise Statement]} and \texttt{[The Start of Code Submitted]} with the exercise statement (i.e., Figure~\ref{fig:ExerciseExamples} left or right) and the student code, respectively.}
\label{fig:prompt_sage}
\end{figure}

\noindent
\textbf{Tools and Requirements.} 
At this point, we hope the reader understands that most exercises depend on Python packages such as \texttt{NumPy}. Indeed, only very simple tasks, such as basic normalization, do not require such external (non-native) libraries. In particular, at the date of this essay, our environment requires only the following Python packages: \texttt{random, itertools, NumPy}, and \texttt{SciPy}. These packages allow us to cover a broad range of fundamental machine learning concepts. As a promising avenue for future exploration, we plan to develop exercises involving more elaborate concepts, including deep learning with \texttt{TensorFlow} and \texttt{PyTorch}.

As a final note, while we employ VPL as our automated programming assessment platform, other platforms can naturally host these exercises, given their simple input-output format, similar to ICPC contests. The platform only needs to provide the essential Python packages mentioned above.
\input{mutex}

\noindent
\textbf{Experience in an Operating Systems Course.} Besides machine learning, we also have two years of experience implementing the input-output assessment strategy in an Operating Systems course. In this course, the exercises cover topics such as \emph{process and memory management} and \emph{parallel concurrency} (including threads, mutexes, and semaphores). For these topics, we primarily use \texttt{C/C++} and \texttt{Assembly x86}. Figure~\ref{fig:Mutex} illustrates an example of a problem involving mutual exclusion with mutex.

Unlike machine learning, some operating systems exercises rely heavily on SAGE. In particular, students can easily obtain the expected output by manipulating the input in ways that diverge from the core learning objectives of the exercise. 
We invite the reader to reflect on how, without the support of SAGE, the exercise in Figure~\ref{fig:Mutex} is susceptible to potential violations\footnote{If you have spent more than five minutes thinking about the problem, you may be overthinking it—the answer is simply to multiply the inputs.}. In this direction, exercises involving pointers require special attention, as students may bypass the intended concepts by using alternative mechanisms, such as registers. These challenges motivate our ongoing efforts to improve SAGE.

\begin{figure*}[!t]
	\centering
	\includegraphics[width=0.3\linewidth]{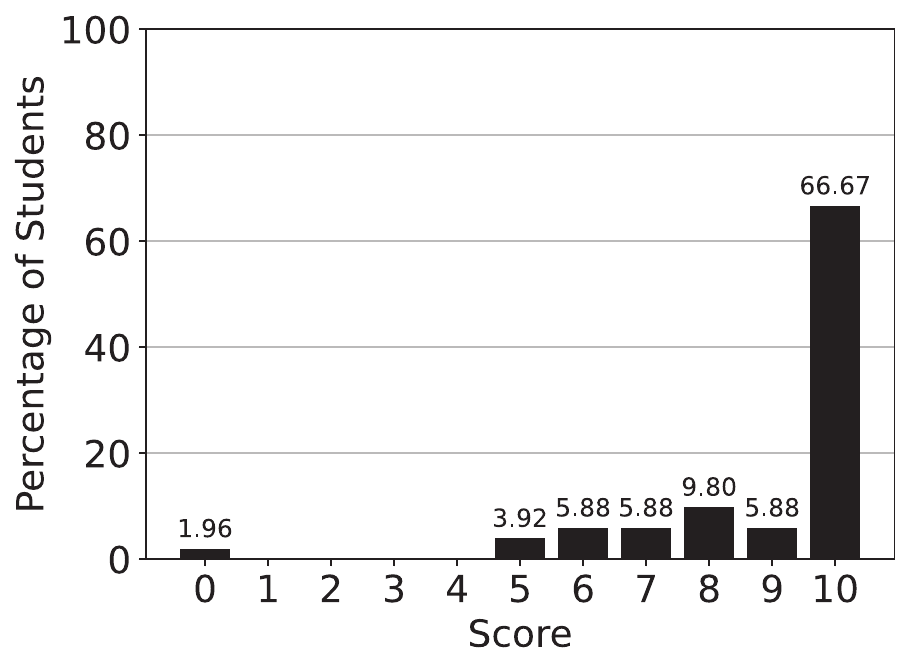}
	\includegraphics[width=0.3\linewidth]{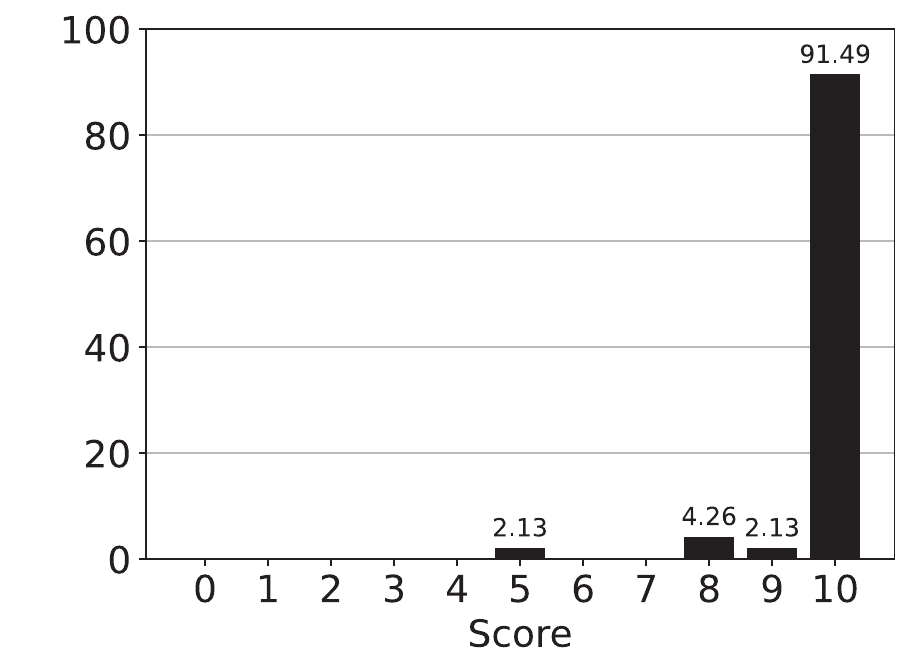}
	\includegraphics[width=0.3\linewidth]{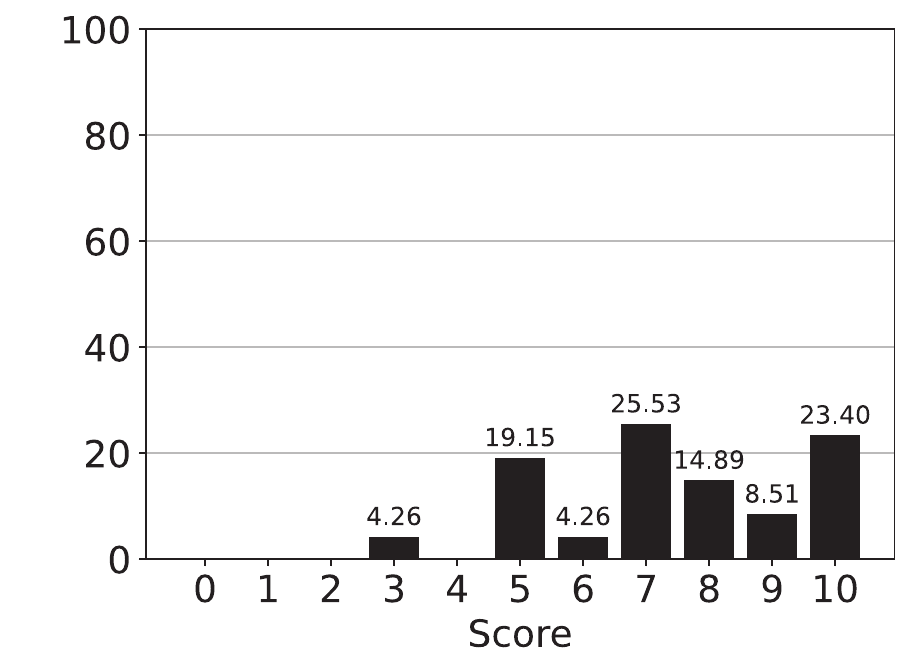}
	\caption{Left. Answer distribution for the question: \emph{Do you believe the automated grading dynamic helped align practical concepts with theory?} Scale: 0 (not helpful) to 10 (completely helpful). Middle. Answer distribution for the question: \emph{How important do you consider the inclusion of input-output examples in the exercise statements?} Scale: 0 (not necessary) to 10 (completely necessary). Right. Answer distribution for the question: \emph{Do you prefer automated or human grading?} Scale: 0 (human only) to 10 (automated only).}
	\label{fig:quantitative}
\end{figure*}

\noindent
\textbf{Student Feedback.}
While our strategy for mapping machine learning problems into input-output assessment tasks may sound particularly interesting to machine learning practitioners and instructors, its primary goal is to bridge the gap between theory and practice and enhance student learning. In Figure~\ref{fig:quantitative}, we summarize the main feedback provided by students across two different courses, including Operating Systems, from 2025 to 2026.

According to Figure~\ref{fig:quantitative} (left), more than 60\% of students fully agree that the exercises improve the integration of theory and practice. In addition, the responses clearly concentrate in the mid-to-high score range, with only one student indicating that the exercises are not helpful. Figure~\ref{fig:quantitative} (middle) reinforces the importance of input-output examples (the tables in Figures~\ref{fig:ExerciseExamples} and~\ref{fig:Mutex}), with responses showing a strong, unimodal peak at the maximum score (10). Indeed, these examples play a vital role in understanding the problem, as in ICPC contests, and the results further support their importance. Finally, Figure~\ref{fig:quantitative} (right) illustrates an interesting behavior when we ask students whether they prefer human or automated grading. From this figure, we observe that responses are more dispersed suggesting greater variability in student opinions. On the one hand, the results in Figure~\ref{fig:quantitative} (right) indicate a clear preference for automated assessment among students. We believe this preference stems from the fast feedback provided by these tools. On the other hand, the concrete scores in the middle range demonstrate that students also appreciate having a human in the loop. Indeed, automated assessment in non-trivial courses can help reallocate instructional time toward higher-impact pedagogical activities, including individualized student tracking.

Overall, Figure~\ref{fig:quantitative} confirms positive student acceptance and suggests a successful experience in integrating theory and practice through automated assessment tools in non-trivial courses we adopt (machine learning and operating systems).

Besides the previous categorical (0-10) feedback, we also asked students to provide open-ended comments about their experience with the activities. Figure~\ref{fig:open-ended_feedback} summarizes student feedback and further supports the effectiveness of our strategy.

\begin{figure}[!h]
{
\scriptsize
\begin{tcolorbox}[width=0.48\textwidth, colback=black!5!white,colframe=black!75!black]
[Student A]
\emph{Increase the number of VPLs. I believe it helped me a lot to learn the concepts, and I would even consider this a suggestion for several subjects, since it is very good to learn concepts in a practical way (even if the exercise is simple) and to reason through developing the solution.}

\vspace{5pt}

[Student B]
\emph{Practical exercises are very good for the continued learning of the subject. Please continue with them.}

\vspace{5pt}

[Student C]
\emph{The platform was excellent for testing theoretical knowledge. In addition to the hands-on exercises provided in the slides, which were great, we also had the opportunity to build code on our own and implement the theory behind it.}

\vspace{5pt}

[Student D]
\emph{Being able to run and write code directly on the platform greatly simplified solving the exercises. I liked this feature a lot.}

\vspace{5pt}

[Student E]
\emph{Automated grading motivates students more to complete the activities, since they do not need to wait long to receive their grade, and therefore know whether they are on the right track or not.}

\vspace{5pt}

[Student F]
\emph{The activities were simple but very good for learning in practice the theory we were seeing in class.}

\vspace{5pt}
[Student G]
\emph{The exercise lists and practical exercises are very good at identifying weaknesses in understanding.}
\end{tcolorbox}
}
\caption{Open-ended student comments about their experience with the activities. We translated the student feedback into English, preserving the original meaning as closely as possible.}
\label{fig:open-ended_feedback}
\end{figure}

\section{Final Remarks}
In this essay, we report a strategy for integrating theory and practice in non-trivial courses, with a focus on machine learning. Overall, our strategy transforms exercises on machine learning concepts into an input-output format. Leveraging this mapping, each exercise admits a unique and deterministic answer and supports the use of automated programming assessment tools. In particular, in this work, we employ Virtual Programming Lab (VPL) for Moodle as our automated programming assessment tool, but the approach is platform-agnostic and thus supports other tools, such as MOJ, Codeforces, and URI Online Judge.
Throughout this essay, we discuss guidelines that can help instructors design effective practical exercises and avoid common issues.
%
We also introduce a large language model–driven inspection mechanism to verify that submissions satisfy the required skill competencies without policy violations, thereby enhancing assessment reliability. We reinforce that this essay \emph{does not introduce an innovation in the use of AI for education}; rather, it introduces an \emph{innovative approach to improving the learning of AI}, particularly, machine learning.

The qualitative and quantitative results indicate strong student acceptance, with students largely agreeing that the exercises effectively integrate theory and practice through automated assessment.
Despite this successful experience and positive feedback, our strategy still has room for improvement. The first is to expand the platform support to additional packages such as \texttt{scikit-learn}. These packages would enable the development of exercises covering more advanced machine learning concepts.
Second, some students point out that the messages for failed test cases (the output does not match the expected value) are uninformative. Addressing this issue is not straightforward, since overly explicit messages, such as \emph{``the expected output is:''}, could make the problem trivial or irrelevant. We highlight that this is not specific to the platform we use (VPL) and also affects ICPC-like contests~\cite{icpc}.

Finally, regarding our strategy for inspecting each submission for potential violations or unauthorized solutions, LLM-based analysis remains susceptible to errors and missed violations. Although our experience with larger models (e.g., 7B-29B parameters) suggested high accuracy on this task, they also impose a high computational cost.

If you found this essay useful, stay tuned to my \href{https://github.com/arturjordao}{GitHub} for updates. Importantly, we echo Donald Knuth’s words from his Claude’s Cycles essay~\cite{Knuth:2026}: \emph{Please work with like-minded researchers as much as you can, but \textbf{without putting me into the loop}!}

\textbf{What's Next?} Thanks to USP’s Information Technology and Technical Support Center (STI), we are currently developing practical activities for deep learning. We are also working to improve SAGE in terms of accuracy and computational efficiency, enabling it to operate at scale in large classes ($>100$ students). An ambitious direction would be to organize an ICPC-style machine learning marathon; we believe this could foster innovation in developing theoretically sound machine learning techniques. 
Perhaps ML experts would even earn one or two balloons.

\section{Acknowledgments}
The author would like to express their deepest gratitude to the students for their constructive feedback. The author would also like to thank Artur Izquerdo for developing the first version of SAGE and Julia Fugita for her valuable contributions to the project. Finally, the author would like to thank the Central de Serviços de TI e Suporte Técnico at USP for providing a strong support environment that enables the development of new features, including deep learning exercises.

\bibliography{refs}
\bibliographystyle{icml2024}
\end{document}

%% file: teaser.tex
\begin{figure*}[t]
{
\tiny
\begin{minipage}[t]{0.48\textwidth}
\begin{tcolorbox}[title=Example of a Support Vector Machines Exercise, colback=black!2!white,colframe=black!75!black]
Write a program that reads two integers, $n$ ($1 \leq n \leq 10^3$) and $m$ ($1 \leq m < n$), representing the number of data points and their dimensionality, respectively, and consequently the dimensionality of the SVM projection matrix (hyperplane) $w$. Next, read the hyperplane $w$ and the bias $b$ of a learned binary SVM model. Both $w$ and $b$ are floating-point values, with each value provided on a separate line. Then, read a dataset $X \in \mathbb{R}^{n \times m}$, also represented using floating-point values. After reading the data, use $w$, $b$, and $X$ to determine and output the number of support vectors. Note: Due to numerical precision issues, use the \texttt{np.isclose(., atol=1e-3)} function provided in the code.

\begin{minipage}[t]{0.4\textwidth}
\centering
\vspace{0pt}
{\renewcommand{\arraystretch}{0.8}
\begin{tabular}{cc}
\hline
Input     & Output \\ \hline
5 2 & 2 \\
-0.9442 -1.4277 & \\
6.4345& \\
0.8730 4.7143& \\
2.1993 2.3519& \\
2.8163 1.0193& \\
1.9263 4.1524& \\
2.84382 3.3265& \\
\hline
\end{tabular}
}
\end{minipage}%
\hspace{20pt}
\begin{minipage}[t]{0.4\textwidth} 
\centering
\vspace{0pt}
{\renewcommand{\arraystretch}{0.8}
\begin{tabular}{cc}
\hline
Input     & Output \\ \hline
6 3 & 3 \\
-0.0682 -0.4551 0.0626 &  \\
0.8990&  \\
1.3868 4.4478 3.5095&  \\
1.5512 -0.6624 2.1757&  \\
1.2313 -0.0328 2.7127&  \\
1.9263 4.1524 1.9520&  \\
1.7373 4.4254 2.4991&  \\
1.2107 -2.3809 0.3648&  \\
\hline
\end{tabular}
}
\end{minipage}%
\hfill
\end{tcolorbox}

\end{minipage}
\hfill
\begin{minipage}[t]{0.48\textwidth}
\begin{tcolorbox}[title=Example of Principal Component Analysis Exercise, colback=white,colframe=black!75!black]
Write a program that reads three integers $s$ $(0 \leq s \leq 1024)$, $n$ $(0 \leq n \leq 10^3)$, and $m$ $(0 \leq m \leq n)$, representing a \texttt{seed} for pseudo-random number generation, the number of samples, and the dimensionality of the data, in this order. Next, read a real number $v \in [0, 1]$. From this data, the program must compute PCA using SVD. Then, the program must output the smallest number of principal components required so that the cumulative explained variance is greater than or equal to $v$. For example, considering $m=5$ and $v=0.8$, if the explained variance proportions are $0.5, 0.2, 0.15, 0.1$, and $0.05$, then the program must output $3$, since the sum of the first three components is $0.85 \geq v$.

Note: Unlike the other problems, here the data is generated pseudo-randomly from the given \texttt{seed}, as illustrated in the code snippet below; therefore, it is important that you do not modify this part. To compute the SVD, use \texttt{np.linalg.svd(., full\_matrices=False)}. To compute the variance, use only $\Sigma^2$. Do not alter the data centering process.

\begin{lstlisting}[language=Python,
basicstyle=\ttfamily,
keywordstyle=\bfseries\color{blue}
]
import numpy as np
import random

np.random.seed(s)
random.seed(s)

X = np.random.randn(n, m)
Xc = X - X.mean(axis=0)
\end{lstlisting}

\begin{minipage}[t]{0.4\textwidth}
\centering
\vspace{0pt}
{\renewcommand{\arraystretch}{0.8}
\begin{tabular}{cc}
\hline
Input     & Output \\ \hline
0 100 10 & 10 \\
0.99& \\
\hline
\end{tabular}
}
\end{minipage}%
\hfill
\begin{minipage}[t]{0.4\textwidth}
\centering
\vspace{0pt}
{\renewcommand{\arraystretch}{0.8}
\begin{tabular}{cc}
\hline
Input     & Output \\ \hline
64 200 12 & 11 \\
0.95& \\
\hline
\end{tabular}
}
\end{minipage}%
\hfill
\end{tcolorbox}
\end{minipage}
}
\caption{Examples of two classical Machine Learning fundamentals mapped to input-output format.}
\label{fig:ExerciseExamples}
\end{figure*}

%% file: mutex.tex
\begin{figure}[!b]
{
\tiny
\begin{minipage}[t]{0.48\textwidth}
\begin{tcolorbox}[title=Example of a Mutual Exclusion Exercise, colback=white,colframe=black]
Write a program that receives two integers, $n$ ($2 \leq n \leq 8$) and $k$ ($10 \leq k \leq 10^3$), indicating the number of threads and the number of iterations, in this order. Implement a mutual exclusion mechanism (mutex) in C++ using inline Assembly with the \texttt{asm volatile} directive. For this purpose, your program must implement the functions \texttt{mutex\_lock()} and \texttt{mutex\_unlock()}. Both functions must use the \texttt{xchg} instruction. Each thread executes the \texttt{worker} function, which simply increments a global variable \texttt{acc} by one in each iteration for the specified number of iterations. The code below illustrates the behavior of the \texttt{worker} function. Note that the program, including its input and output, follows the same format as the activity presented in class.

\begin{lstlisting}[language=C++, basicstyle=\ttfamily, % Fonte monoespaçada
keywordstyle=\bfseries\color{blue}, % Palavras-chave em azul e negrito
commentstyle=\itshape\color{gray},  % Comentários em itálico e cinza
stringstyle=\color{red},  % Strings em vermelho
%numbers=left,        % Numeração das linhas à esquerda
numberstyle=\tiny,   % Tamanho pequeno para os números das linhas
stepnumber=1,        % Numera todas as linhas
breaklines=true      % Quebra de linha automática
]
void* worker(void *thread_id){
   for (int i=0; i<k; i++){
     mutex_lock();
     acc = acc + 1;
     mutex_unlock();
   }
   return nullptr;
}
\end{lstlisting}

Note: (i) You may use only the \texttt{pthread\_create} and \texttt{pthread\_join} functions from the pthread library. (ii) Your program must implement a mutual exclusion mechanism using the \texttt{xchg} instruction, avoiding the trivial and incorrect solution of multiplying \texttt{n\_threads} by \texttt{iterations}. (iii) Your code must not modify the \texttt{worker} function. (iv) Your program must execute exclusively using threads, with each thread executing the \texttt{worker} function.

\begin{minipage}[t]{0.32\textwidth} 
\centering
\vspace{0pt}
\begin{tabular}{cc}
\hline
Input     & Output \\ \hline
2 10000    & 20000    \\ 
\hline
\end{tabular}
\end{minipage}%
\hfill
\begin{minipage}[t]{0.32\textwidth} 
\centering
\vspace{0pt}
\begin{tabular}{cc}
\hline
Input     & Output \\ \hline
3 7240    & 21720    \\ 
\hline
\end{tabular}
\end{minipage}%
\hfill
\begin{minipage}[t]{0.32\textwidth} 
\centering
\vspace{0pt}
\begin{tabular}{cc}
\hline
Input    & Output \\ \hline
6 4000100        &    24000600   \\ \hline
\end{tabular}
\end{minipage}
\hfill
\end{tcolorbox}
\end{minipage}
}
\caption{Example of an important Operating Systems concept (mutual exclusion) mapped to an input-output format.}
\label{fig:Mutex}
\end{figure}